\pdfoutput=1

\documentclass[11pt]{article}
\usepackage{times}
\usepackage{latexsym}
\usepackage{graphicx}
\usepackage{url}
\usepackage{algorithm}
\usepackage{algpseudocode}
\usepackage{amsmath}
\usepackage{amssymb}
\usepackage{multirow}
\usepackage{arydshln}
\usepackage[final]{acl}

\usepackage{times}
\usepackage{latexsym}

\usepackage[T1]{fontenc}

\usepackage[utf8]{inputenc}

\usepackage{microtype}

\title{UniXcoder: Unified Cross-Modal Pre-training for Code Representation}

\author{Daya Guo$^1$\thanks{\ \ \ Work done while this author was an intern at Microsoft Research. Contact: Daya Guo (guody5@mail2.sysu.edu.cn).} , Shuai Lu$^2$, Nan Duan$^2$, Yanlin Wang$^2$, Ming Zhou$^3$, and Jian Yin$^1$\\
$^1$ School of Computer Science and Engineering, Sun Yat-sen University.\\
	Guangdong Key Laboratory of Big Data Analysis and Processing, Guangzhou, P.R.China\\
	$^2$ Microsoft Research Asia, Beijing, China\\
	$^3$ Langboat Technology, Beijing, China \\
}

\begin{document}
\maketitle
\begin{abstract}
Pre-trained models for programming languages have recently demonstrated great success on code intelligence.
To support both code-related understanding and generation tasks, recent works attempt to pre-train unified encoder-decoder models. 
However, 
such encoder-decoder framework is sub-optimal for auto-regressive tasks, especially code completion that requires a decoder-only manner for efficient inference.
In this paper, we present UniXcoder, a unified cross-modal pre-trained model for programming language. The model utilizes mask attention matrices with prefix adapters to control the behavior of the model and leverages cross-modal contents like AST and code comment to enhance code representation. 
To encode AST that is represented as a tree in parallel, we propose a one-to-one mapping method to transform AST in a sequence structure that retains all structural information from the tree.
Furthermore, we propose to utilize multi-modal contents to learn representation of code fragment with contrastive learning, and then align representations among programming languages using a cross-modal generation task. 
We evaluate UniXcoder on five code-related tasks over nine datasets. To further evaluate the performance of code fragment representation, we also construct a dataset for a new task, called zero-shot code-to-code search. Results show that our model achieves state-of-the-art performance on most tasks and analysis reveals that comment and  AST can both enhance UniXcoder. 


\end{abstract}

\section{Introduction}
Pre-trained models such as GPT \citep{radford2018improving} and BERT \citep{devlin2018bert} have substantially advanced the state of the art across numerous natural language processing (NLP) tasks.  These pre-trained models are  pre-trained on large amounts of text data with self-supervised objectives, and can be fine-tuned to adapt to downstream tasks.
Inspired by the success of pre-trained models in NLP, pre-trained models for programming languages (PL) \citep{kanade2019pre,feng2020codebert,svyatkovskiy2020intellicode}
have been proposed to promote the development of code intelligence.
\citet{svyatkovskiy2020intellicode} proposes GPT-C that employs a left-to-right Transformer \cite{vaswani2017attention} to support generation tasks such as code completion, but the unidirectional framework is sub-optimal for understanding tasks. 
In contrast,  other works \cite{kanade2019pre,feng2020codebert} pre-train a bidirectional Transformer encoder on source code, which significantly improves the performance of code-related understanding tasks. However, its bidirectionality nature requires an additional decoder when applied to generation tasks, where this decoder initializes from scratch and cannot benefit from the pre-training.


In this work, we present UniXcoder, a unified cross-modal pre-trained model for programming languages to support both code-related understanding and generation tasks. 
UniXcoder is based on a multi-layer Transformer and follows \citet{dong2019unified} to utilize mask attention matrices with prefix adapters to control the access to context for each token.
Compared with current unified encoder-decoder models \cite{ahmad2021unified,wang2021codet5} on code intelligence, UniXcoder can be better applied to auto-regressive tasks such as code completion that requires a decoder-only manner to perform efficient inference in practice. 
Instead of taking code as the only input, we also consider multi-modal contents like code comment and abstract syntax tree (AST) to enhance code representation. Generally, user-written code comments provide crucial semantic information about source code like \textit{``Sort a given list''} and AST contains rich syntax information like types of statements and nested relationship among them, which helps the model better understand source code. To encode AST that is represented as a tree in parallel, we propose a one-to-one mapping method to transform AST in a sequence structure that retains all information of the tree and then the sequence can be used as the input to enhance code representation.

We pre-train UniXcoder using three types of language modeling tasks: masked language modeling \cite{devlin2018bert}, unidirectional language modeling \cite{radford2018improving} and denoising objective \cite{raffel2019exploring}, which can enable the model to support various types of downstream tasks. Furthermore, we introduce two pre-training tasks to learn a embedding that can represent semantics of a code fragment. One is multi-modal contrastive learning that leverages AST to enhance semantics of code fragment embeddings, and the other is cross-modal generation that utilizes code comment to align embeddings among programming languages. 

We evaluate UniXcoder on five tasks over nine public datasets, including two understanding tasks: clone detection and code search, two generation tasks: code summarization and code generation, and an auto-regressive task: code completion.
To further test code fragment embeddings, we propose a new task, called zero-shot code-to-code search, and construct a new dataset from CodeNet corpus \cite{puri2021project} for this task. 
Experimental results show that our model achieves state-of-the-art performance on most tasks.
Further analysis reveals that AST and code comment can both enhance UniXcoder to better capture code semantics.

In summary, the contributions of this paper are: (1) We propose a unified cross-modal pre-trained model that leverages multi-modal contents, i.e. code comment and AST, to support code-related understanding, generation tasks and auto-regressive tasks. (2) We propose a one-to-one mapping function that converts AST into a sequence that retains all information of AST and can be encoded with source code and comment in parallel.  (3) We further propose to utilize code comment to learn code fragment representation and construct a new dataset for zero-shot code-code search to evaluate the quality of code fragment representation. (4) Experimental results show that UniXcoder provides significant improvement on most downstream tasks.\footnote{All the codes and data are available at \url{https://github.com/microsoft/CodeBERT}.}

\section{Related Works}
With the great success of pre-training in natural language (NL)  processing \citep{devlin2018bert,lewis2019bart,raffel2019exploring,brown2020language}, pre-trained models for programming languages have been proposed to promote the development of code intelligence.
These pre-trained models can be generally divided into three categories: encoder-only, decoder-only, and encoder-decoder models.

Encode-only models \cite{kanade2019pre,buratti2020exploring,feng2020codebert,guo2020graphcodebert,wang2022syncobert} pre-train a bidirectional Transformer in which each token can attend to each other.
\citet{kanade2019pre} pre-train CuBERT on a corpus of Python source codes by masked language modeling and next sentence prediction objectives. CodeBERT\cite{feng2020codebert} is pre-trained on NL-PL pairs in six programming languages with a new pre-training task, namely replace token detection. GraphCodeBERT \cite{guo2020graphcodebert} leverages data flow to enhance code representation, while \textsc{SynCoBERT} \cite{wang2022syncobert} incorporates abstract syntax tree by AST edge prediction and contrastive learning.
However, encoder-only models require an additional decoder for generation tasks, where this decoder initializes from scratch and cannot benefit from the pre-training.

As for decoder-only pre-trained models, \citet{svyatkovskiy2020intellicode} and \citet{lu2021codexglue} respectively propose GPT-C and CodeGPT, which are both pre-trained using unidirectional language modeling that only allows tokens to attend the previous tokens and itself to predict the next token. 
Decoder-only models are good at auto-regressive tasks like code completion, but the unidirectional framework is sub-optimal for understanding tasks.

Some recent works explore encoder-decoder models to support both understanding and generation tasks. PLBART \citep{ahmad2021unified} is based on the BART \citep{lewis2019bart} architecture and pre-trained on NL and PL corpus using denoising objectives. CodeT5 \citep{wang2021codet5} adapts the T5 \citep{raffel2019exploring} model that considers the crucial token type information from identifiers and allow for multi-task learning on downstream tasks.
TreeBERT \citep{jiang2021treebert} follows the encoder-decoder transformer framework but utilizes the tree structural information by modeling AST paths.

Different from current unified models, UniXcoder is based on a multi-layer Transformer and utilizes mask attention matrices with prefix adapters to control the behavior of the model for supporting both understanding and generation tasks.
Compared with the encoder-decoder architecture, UniXcoder can be better applied to auto-regressive tasks like code completion that is widely used in IDEs, since the task requires a decoder-only manner to perform efficient inference in practice. \citet{liu2020multi} also pre-train a similar model CugLM with multi-task learning, but they only focus on code completion rather than various tasks. Besides, we incorporate syntax information from AST by a one-to-one mapping function that converts an AST into a sequence to enhance code representation. Different from previous pre-trained models that utilize AST, the mapping function retains all structural information from AST and does not require additional pre-training tasks (such as edge prediction) to implicitly learn the AST structure.

\section{UniXcoder}
In this section, we describe UniXcoder, a unified cross-modal pre-trained model that leverages multi-modal data (i.e. code comment and AST) to pre-train code representation. The model is based on Transformer and utilizes mask attention matrices \cite{dong2019unified} with prefix adapters to control the behavior of the model. In the following, we first introduce how to unify multi-modal data as the input of UniXcoder (\S \ref{sec:input}), and then the model architecture (\S \ref{sec:model}) and pre-training tasks (\S \ref{sec:tasks}). 

\subsection{Input Representation}
\label{sec:input}
\begin{figure}
    \centering
    \includegraphics[width=0.98\linewidth]{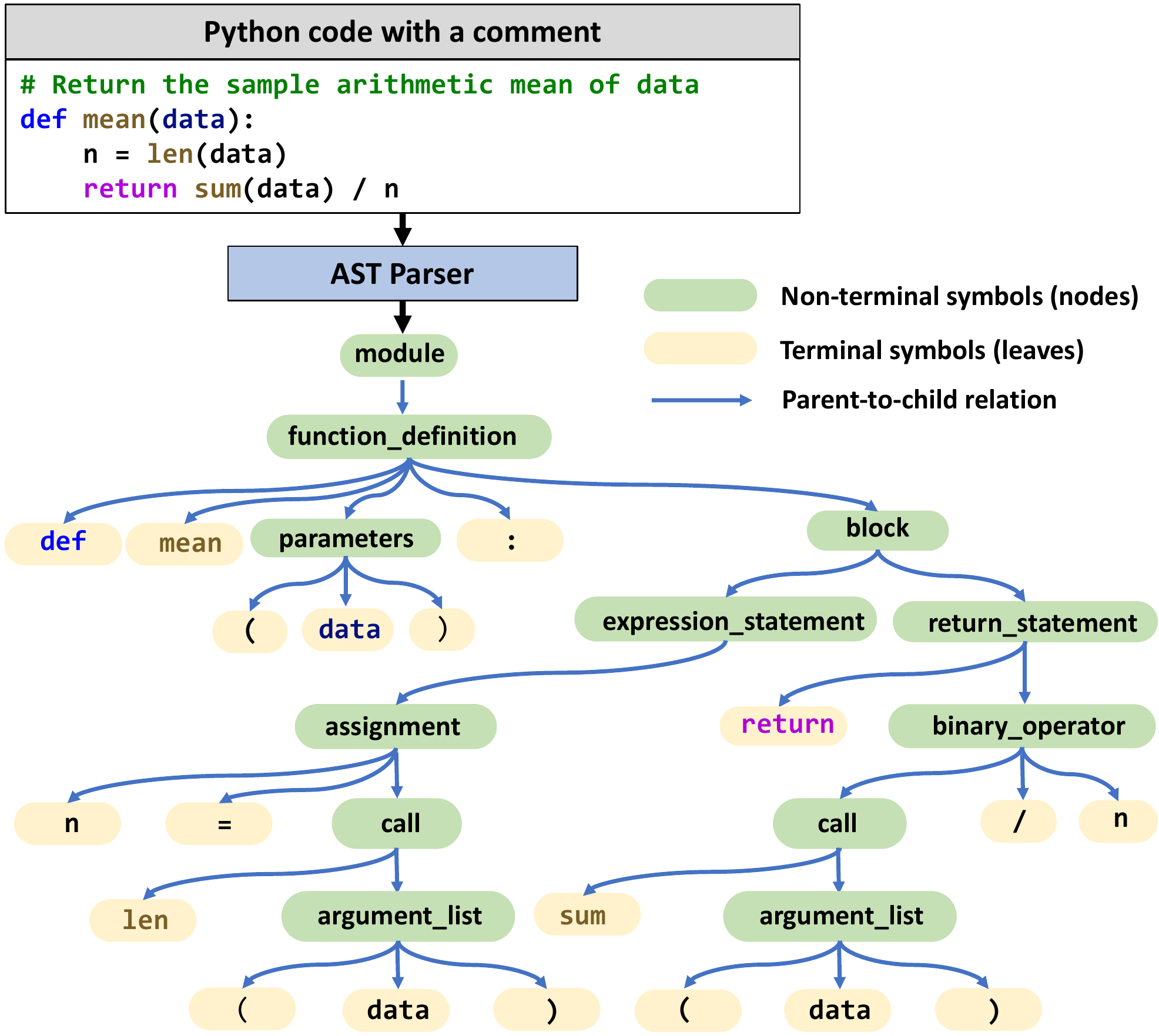}
    \caption{A Python code with its comment and AST.}
    \label{fig:AST}
\end{figure}

We give an example of a python code with its comment and AST in Figure \ref{fig:AST}. From the figure, we can see that the comment \textit{``Return the sample arithmetic mean of data''}  highly describes the function of the source code, which provides crucial semantic information about the source code. 
Besides, AST provides rich syntax information, for example, the subtree \textit{``parameters $\rightarrow$ (data)''} indicates the type (i.e., \texttt{parameters}) of the term \texttt{(data)} in the function definition. 
Both of them can be used as additional knowledge to enhance code representation in pre-trained models. However, AST is usually expressed as a tree and cannot be used directly as input to Transformer.
In order to encode AST in parallel with code comments, we propose a one-to-one mapping function $\mathcal{F}$, described in Algorithm \ref{alg:mapping}, to transform an AST into a sequence that retains all structural information.

\begin{algorithm}[htb]
	\caption{AST Mapping Function $\mathcal{F}$}
	\label{alg:mapping}
	\begin{algorithmic}[1]
		\Require \
		The root node $root$ of AST
		\Ensure \
		A flattened token sequence
		\Function {$\mathcal{F}$}{$root$}  
		\State $seq =$ an empty list
		\State $name =$ the name of $root$
		\If {$root$ is a leaf}
		\State $seq.append(name)$
	    \Else 
	    \State $seq.append(name::left)$
	    \For{$child$ in children of $root$}
	    \State $seq.extend(\mathcal{F}(child))$
	    \EndFor
	    \State $seq.append(name::right)$
	    \EndIf
	    \EndFunction
	\end{algorithmic}
	
\end{algorithm}

Specially, given a root node $root$ of AST, the algorithm recursively applies the same function $\mathcal{F}$ to its children and then add its name with two special suffixes (i.e. $left$ and $right$, respectively) on both sides (line 6-11 of Algorithm  \ref{alg:mapping}). If the root node is a leaf, we directly produce its name (line 4-5). Taking \textit{``parameters $\rightarrow$ (data)''} as an example, the mapping function $\mathcal{F}$ transforms the subtree to \textit{``<parameters,left> ( data ) <parameters,right>''}.

\begin{figure*}[h]
    \centering
    \includegraphics[width=0.95\linewidth]{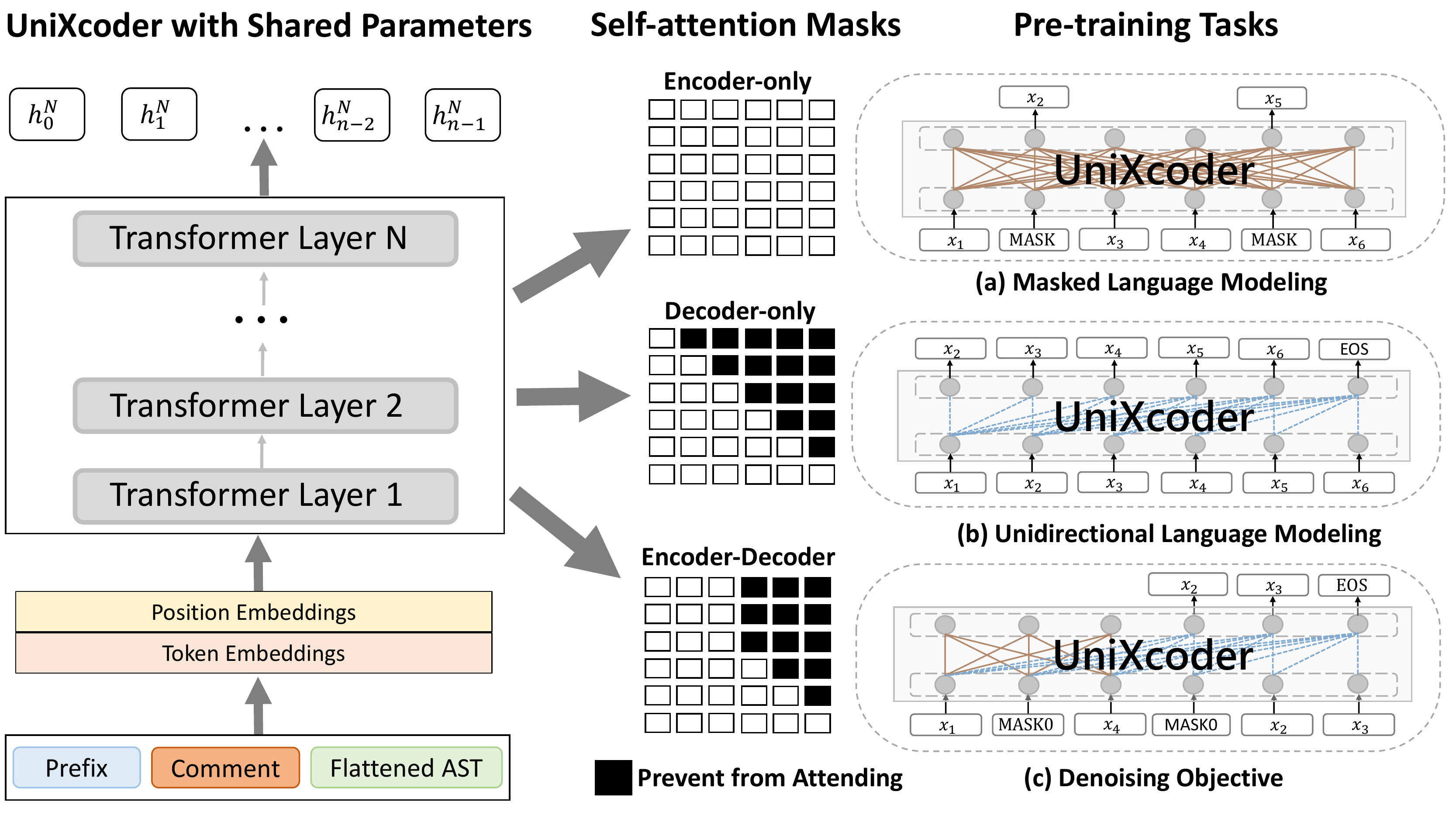}
    \caption{Model architecture of UniXcoder. The model takes comment and flattened AST as the input (more specific input examples can be found  in Figure \ref{fig:contras}). Model parameters are shared in different modes. We use different self-attention masks to control the behavior of the model and use various tasks to pre-train the model, including masked language modeling, unidirectional language modeling, and denoising objective. 
}  
    \label{fig:model}
\end{figure*}

There can be various ways to transform a tree to a sequence of tokens, e.g. pre-order traversal. However, a particular transformation should be a one-to-one mapping function. Otherwise, the mapping may confuse a tree with another structure. Our mapping function $\mathcal{F}$ satisfies this requirement (see Appendix \ref{sec:proof} for a proof). Finally, given a source code $C$, we take its comment $W=\{w_0,w_1,...,w_{m-1}\}$ and the flattened AST token sequence   $\mathcal{F}(\mathcal{T}(C))=\{c_0,c_1,...,c_{k-1}\}$ as input, where $\mathcal{T}(C)$ is the root of the AST of the code. 
For input format, we concatenate them with a prefix as an input sequence, as shown at the bottom of Figure \ref{fig:model}, where the prefix represents the work mode of the model and will be discussed next.


\subsection{Model Architecture}
\label{sec:model}
Figure \ref{fig:model} shows the model architecture of UniXcoder. The model applies N transformer layers over code comment and flattened AST with a prefix to produce hidden states $H^N=\{h_0^N,h_1^N,...,h_{n-1}^N\}$, where the prefix $p\in \{[Enc],[Dec],[E2D]\}$ indicates the behavior of the model, e.g. $[E2D]$ means that UniXcoder works as a encoder-decoder model. 
Each transformer layer contains an architecturally identical transformer that uses a multi-headed self-attention operation \citep{vaswani2017attention} followed by a feed forward layer over the output of the previous layer. 
For the $l$-th transformer layer, the output of the multi-headed self-attention is computed via:
\begin{equation}
Q=H^{l-1}W^{Q},K=H^{l-1}W^{K},V=H^{l-1}W^{V}
\end{equation}
\begin{equation}\label{eq:score}
head=\rm{softmax}(\frac{QK^T}{\sqrt{d_k}}+M)V
\end{equation}
where 
previous layer's output $H^{l-1}\in\mathbb{R}^{n \times d_h}$ is linearly mapped to a triplet of queries, keys and values 
respectively. $d_k$ is the dimension of a head, and $M\in\mathbb{R}^{n \times n}$ is a mask matrix to control the  context a token can attend to when computing its contextual representation, as shown in the middle of Figure \ref{fig:model}. If the $i$-th token is allowed to attend to the $j$-th token, then $M_{ij}$ is set to 0 otherwise  $-\infty$. 

For encoder-only mode, we add a special token $[Enc]$ as the prefix in front of the input and set all elements of the mask matrix as 0 to allow all tokens attend to each other.
For decoder-only mode, a prefix $[Dec]$ is used and the upper triangular part of the mask is set to $-\infty$ to indicate that each token can only attend to itself and previous tokens.
For encoder-decoder mode, tokens in the source input are allowed to attend to each other, while tokens in the target input only attend to itself and previous tokens in both source and target inputs. We use the $[E2D]$ prefix to indicate that UniXcoder works as an encoder-decoder model. During the pre-training phase, model parameters are shared in different modes and optimized with several objectives to support various types of downstream tasks.

\subsection{Pre-training Tasks}
\label{sec:tasks}
\begin{figure}
    \centering
    \includegraphics[width=\linewidth]{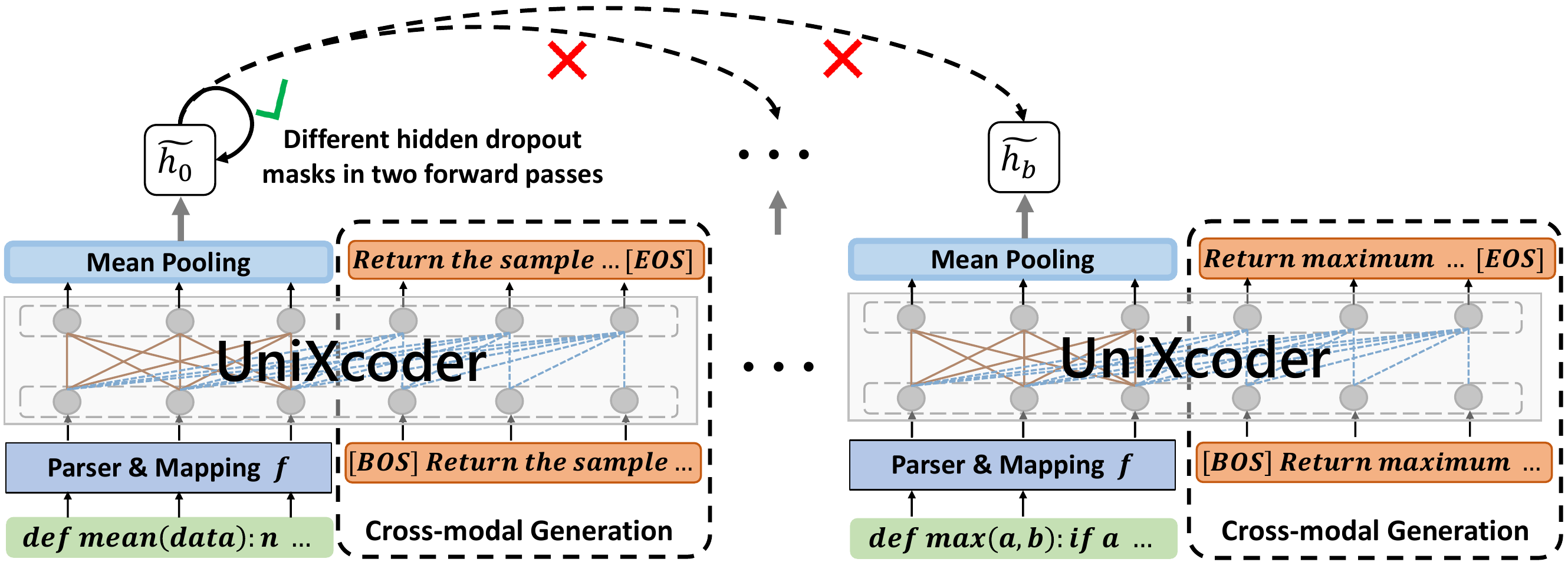}
    \caption{Code fragment representation learning.}
    \label{fig:contras}
\end{figure}

We describe the pre-training tasks used in UniXcoder in this section. As shown on the right side of Figure \ref{fig:model}, we first pre-train UniXcoder using three tasks, including masked language modeling \cite{devlin2018bert}, unidirectional language modeling \cite{radford2018improving} and denoising objective \cite{raffel2019exploring}. These tasks are designed for different modes,  enabling UniXcoder to support various types of code-related downstream tasks. We then propose to utilize multi-modal data to learn code fragment embeddings through contrastive learning with cross-modal generation, as shown in Figure \ref{fig:contras}.

\paragraph{Masked Language Modeling}
For encoder-only mode, we follow \citet{devlin2018bert} to apply masked language modeling (MLM) pre-training task. 
Specially, we sample 15\% of the tokens $S_{m}$ from the input sequence, and then replace 80\% (10\%) of them with a [MASK] (random) token and leave another 10\% of them unchanged.
The task is to predict original tokens of masked tokens based on their bidirectional contextual tokens, as illustrated in Figure \ref{fig:model} (a).
In particular, the model can leverage semantic information from comment and syntax information from AST to infer masked code tokens, which encourages the model to learn code representations from different knowledge resources. The objective is calculated as Equation \ref{eq:mlm}, where $X^{mask}$ is the masked input sequence.
\begin{equation}\label{eq:mlm}
loss_{MLM}=-\sum_{x_i\in S_{m}} log p(x_i|X^{mask})
\end{equation}
\paragraph{Unidirectional Language Modeling}
We use unidirectional language modeling (ULM) pre-training task to pre-train decoder-only mode for supporting auto-regressive tasks like code completion, as shown in Figure \ref{fig:model} (b). The task predicts the next token $x_i$ one by one conditioned on previous tokens and itself $\{x_0,x_1,..,x_{i-1}\}$, which can be done using a triangular matrix for attention mask. 
\begin{equation}\label{eq:ulm}
loss_{ULM}=-\sum_{i=0}^{n-1} log p(x_i|x_{t<i})
\end{equation}

\paragraph{Denoising Objective} 
{\textbf D}e{\textbf N}oi{\textbf S}ing (DNS) pre-training objective has been shown to be quite effective for encoder-decoder models like BART \citep{lewis2019bart} and T5 \citep{raffel2019exploring} in NLP. The task randomly masks spans with arbitrary lengths and then generates these masked spans in encoder-decoder mode. To better support generation tasks like code summarization, we utilize similar denoising objective  as T5 for encoder-decoder mode, as illustrated in Figure \ref{fig:model} (c). 
Specially, we first split the input sequence into $max(\lfloor  \frac{n\times r}{l} \rfloor,1)$ chunks and then randomly mask a span of from 1 to $2l$-$1$ tokens for each chunk, where $n$ is the length of the input, $r$ is corruption rate and $l$ is the average length of masked spans. We set corruption rate as 15\% and the average length as 5, respectively. The concatenation $\{y_0,y_1,...,y_{n-1}\}$ of all masked spans with special tokens $[MASK_k]$ in front of the $k$-th span will be used as the output:
\begin{equation}\label{eq:score}
loss_{DNS}=-\sum_{i=0}^{n-1} log p(y_i|X^{mask},y_{t<i})
\end{equation}

\paragraph{Code Fragment Representation Learning}

In addition to the above three pre-training tasks designed for different modes, we propose to utilize multi-modal data to learn semantic embedding $\widetilde{h}_i$ of a code fragment $C_i$. As shown in Figure \ref{fig:contras}, we first use UniXcoder to encode a mapped AST sequence and then apply a mean pooling layer over the hidden states of the source input to obtain semantic embedding $\widetilde{h}_i$. In order to learn the semantic embedding, we propose two pre-training tasks. One is multi-modal contrastive learning  (MCL), and another is cross-modal generation  (CMG). 

For multi-modal contrastive learning, we follow \citet{gao2021simcse} to forward the same input using different hidden dropout mask as a positive example $\widetilde{h}_{i}^+$ and use other representations in the same batch as negative examples. The loss is calculated as Equation \ref{eq:mcl}, where $b$ is batch size, $\tau$ is a temperature hyperparameter, and $cos(\cdot ,\cdot)$ is the cosine similarity between two vectors.  
\begin{equation}\label{eq:mcl}
loss_{MCL}=-\sum_{i=0}^{b-1} log\frac{e^{cos(\widetilde{h}_i,\widetilde{h}_{i}^+)/\tau}}{\sum_{j=0}^{b-1}e^{cos(\widetilde{h}_i,\widetilde{h}_{j}^+)/\tau}}
\end{equation}

For cross-modal generation, we ask the model to 
generate its comment $W=\{w_0,w_1,...,w_{m-1}\}$.  The comment describes the function of the code, which can help the model not only understand the code semantics but align representations among different programming languages by a unified natural language description as a fulcrum. Since the generation of the comment is conditioned on the code, it will force the model to fuse semantic information from the comment into the hidden states of the code. The loss is calculated as Equation \ref{eq:cmg}, where $X$ is the flattened AST token sequence.
\begin{equation}\label{eq:cmg}
loss_{CMG}=-\sum_{i=0}^{m-1} log p(w_i|X,w_{t<i})
\end{equation}

In order to learn the semantic embedding of natural language, we randomly exchange the source input and the target input with a probability of 50\%.

Considering that explicitly adding AST in downstream tasks will introduce extra costs like parsing time and increasing input length (70\% longer input length after tokenization), we implicitly learn knowledge from AST by pre-training and only keep leaves of AST (i.e. source code) in the fine-tuning phase. This gap can be alleviated by randomly drop all non-terminal symbols of AST with a probability of 50\% in the pre-training phase.  More details about pre-training dataset and settings can be found in the Appendix \ref{appedix:pre-training-setting}.
\begin{table*}[h]
\small
	\centering
	\begin{tabular}{l|c|c|c|c|c|c|c}
		\hline
		\multirow{3}{*}{Model} & \multicolumn{4}{c|}{Clone Detection} & \multicolumn{3}{c}{Code Search}\\
		\cline{2-8}
		&POJ-104 &\multicolumn{3}{c|}{BigCloneBench} & CosQA &AdvTest &CSN\\
		\cline{2-8}
		& MAP@R &Recall&Precision& F1-score& \multicolumn{3}{c}{MRR} \\
		\hline
		RoBERTa & 76.67& \bf{95.1}&87.8&91.3& 60.3 & 18.3 & 61.7\\
		CodeBERT & 82.67&94.7&93.4& 94.1& 65.7 & 27.2& 69.3\\
		GraphCodeBERT & 85.16&94.8&95.2& 95.0& 68.4 & 35.2& 71.3\\
        \textsc{SynCoBERT} & 88.24&-& -&-& - & 38.3& 74.0\\
        \hline
		PLBART & 86.27 &94.8&92.5& 93.6& 65.0 & 34.7 & 68.5\\        
		CodeT5-base & 88.65&94.8&94.7 & 95.0& 67.8 & 39.3& 71.5 \\
        \hline
        UniXcoder &\bf{90.52}&92.9&\bf{97.6}&\bf{95.2}&\bf{70.1}&\bf{41.3}&\bf{74.4} \\ 
        \hdashline
        \ -w/o contras &87.83&94.9&94.9&94.9&69.2&40.8&73.6  \\
        \ -w/o cross-gen &90.51&94.8&95.6&95.2&69.4&40.1&74.0  \\
        \hdashline
        \ -w/o comment &87.05&93.6&96.2&94.9&67.9&40.7&72.6\\
        \ -w/o AST &88.74&92.9&97.2&95.0&68.7&40.3&74.2\\
        \hdashline
        \ -using BFS &89.44&93.4&96.7&95.0&69.3&40.1&74.1\\
        \ -using DFS &89.74&94.7&94.6&94.7&69.0&40.2&74.2\\        
		\hline
	\end{tabular}
	\caption{Results on understanding tasks. \textbf{contras} is contrastive learning, \textbf{cross-gen} indicates cross-modal generation, and \textbf{BFS} (\textbf{DFS}) means that our mapping function is replaced by breath-first (deep-first) search algorithm.}
	\label{table:understand_task}
\end{table*}
\section{Experiments}
We evaluate UniXcoder on five tasks over nine public datasets, including two understanding tasks (\S \ref{sec:understanding}),
two generation tasks (\S \ref{sec:generation})
and an auto-regressive task (\S \ref{sec:completion}).
To further evaluate the performance of code fragment embeddings, we also propose a new task called zero-shot code-to-code search (\S \ref{sec:zero-shot}). More details about datasets and fine-tuning can be found in the Appendix \ref{appedix:fine-tuning-setting}.

\subsection{Baselines}
We compare UniXcoder with state-of-the-art pre-trained models, including \textbf{encoder-only}, \textbf{decoder-only} and  \textbf{encoder-decoder} models. 

For \textbf{encoder-only} models, we consider Roberta \cite{liu2019roberta} pre-trained on text corpus with MLM, CodeBERT \cite{feng2020codebert}  pre-trained on NL-PL pairs using both MLM and replaced token detection, GraphCodeBERT \cite{guo2020graphcodebert} that leverages data flow to enhance code representation, and \textsc{SynCoBERT} that incorporates AST by edge prediction and contrastive learning. 

For \textbf{decoder-only} models, we consider GPT-2 \cite{radford2019language} and CodeGPT \cite{lu2021codexglue}, where the former one is pre-trained on text corpus and the latter one is pre-trained on CodeSearchNet dataset. Both use ULM as the objective. 

For \textbf{encoder-decoder} models, we mainly compare the  current unified models PLBART \cite{ahmad2021unified} and CodeT5 \cite{wang2021codet5}. PLBART is based on BART and pre-trained on 470M Python and 210M Java functions, and 47M NL posts from StackOverflow using denoising objective. 
CodeT5, adapted from T5, considers the crucial token type information from identifiers and allows multi-task learning on downstream tasks. 

\subsection{Understanding Tasks}
\label{sec:understanding}


\paragraph{Clone Detection}
The task is to measure the similarity between two code fragments. We conduct experiments on POJ-104 \cite{mou2016convolutional} and BigCloneBench \citep{svajlenko2014towards} datasets. The first dataset is to predict whether two codes have the same semantics and uses F1-score as the evaluation metric, while the second aims to retrieve semantically similar codes given a code as the query with the Mean Average Precision (MAP) as the metric. 

\paragraph{Code Search} 
The task aims to find the most relevant code from a collection of candidates given a natural language query. We conduct experiments on three datasets, namely CSN \cite{guo2020graphcodebert}, AdvTest \cite{lu2021codexglue} and CosQA \cite{huang2021cosqa}. CSN dataset is constructed from CodeSearchNet dataset of six programming languages, and low-quality queries are filtered by handcrafted rules. AdvTest normalizes python function and variable names to better test the understanding and generalization capabilities of models. The code base of CosQA is also from CodeSearchNet 
corpus but queries come from the search logs of Microsoft Bing search engine. We use Mean Reciprocal Rank (MRR) evaluation metric for the task.

\paragraph{Results} 
The results are shown in Table \ref{table:understand_task}. Compared with encoder-only pre-trained models (i.e. the first group) and encoder-decoder models (i.e. the second group), 
UniXcoder outperforms them and achieves state-of-the-art performance on two tasks on all five datasets. By comparing with the results of ablation studies in the last six rows, we can see that the improvement mainly comes from  contrastive learning and the use of multi-modality. 


\subsection{Generation Tasks}
\label{sec:generation}
\paragraph{Code Summarization}
The task aims to generate an NL summary of a code snippet. We use the dataset provided by the CodeXGLUE team \cite{lu2021codexglue} for this task. We use the smoothed BLEU-4 \cite{lin2004orange} as the evaluation metric and report overall score of six PLs, including Ruby, JavaScript, Go, Python, Java, and PHP.

\paragraph{Code Generation}  The task is to generate a code snippet based on an NL description. we use CONCODE \cite{iyer2018mapping} dataset, where the input consists of an NL description and code environments. For this task, we use exact match (EM) and BLEU-4 as evaluation metrics.

\begin{table}[h]
\small
	\centering
	\begin{tabular}{l|c|ccc}
		\hline
		\multirow{2}{*}{Model} & Summarization  & \multicolumn{2}{c}{Generation} \\
		\cline{2-4}
		&BLEU-4 & EM&BLEU-4\\
		\hline
	    RoBERTa&16.57 & - &-\\
	    CodeBERT&17.83 & - &-\\
	    \hline
	    GPT-2&- & 17.35 &25.37\\
	    CodeGPT&-&20.10&32.79\\
	    \hline
	    PLBART&18.32 & 18.75 &36.69\\
	    CodeT5-small&19.14 & 21.55 &38.13\\
	    CodeT5-base&\bf{19.55} & 22.30 &\bf{40.73}\\
		\hline
        UniXcoder&19.30 & \bf{22.60} &38.23\\ 
        \hdashline
        \ -w/o contras &19.20 & 22.10 &37.69\\ 
        \ -w/o cross-gen &19.27 & 22.20 &35.93\\ 
        \hdashline
        \ -w/o comment &18.97 & 21.45 &37.15\\ 
        \ -w/o AST &19.33 & 22.60 &38.52\\ 
        \hdashline
        \ -using BFS &19.24 & 21.75 &38.21\\ 
        \ -using DFS &19.25 & 22.10 &38.06\\         
		\hline
	\end{tabular}
	\caption{Results on two generation tasks, including code summarization and code generation.}
	\label{table:generation_task}
\end{table}

\paragraph{Results} From Table \ref{table:generation_task}, UniXcoder achieves comparable performance on generation tasks compared with CodeT5-base and brings a 0.3\% improvement in code generation accuracy. However, UniXcoder has slightly worse BLEU-4 scores on both code summarization and generation tasks. The main reasons may come from two aspects. One is the amount of NL-PL pairs in the pre-training data. As shown in the ablation study (see \textbf{w/o comment}) in the table, NL-PL pairs bring significant improvement on two tasks. \citet{wang2021codet5} collect 50\% more NL-PL pairs from Github to pre-train CodeT5. Since the collected data is not public, we cannot use it to pre-train UniXcoder for fair comparison. Anothor reason is the model size.
CodeT5-base uses a 12-layer encoder and a 12-layer decoder, which is twice larger than other baselines and UniXcoder. Therefore, we also list the results of CodeT5-small using  a 6-layer encoder and a 6-layer decoder. We can see that UniXcoder outperforms CodeT5-small.

\subsection{Code Completion}
\label{sec:completion}
We use \textbf{PY150} \cite{raychev2016probabilistic} and \textbf{Github Java Corpus} \cite{allamanis2013mining} datasets in CodeXGLUE \cite{lu2021codexglue} for line-level code completion tasks. The task entails the completion of a whole-line of code, and is evaluated using exact match accuracy and Levenshtein edit similarity \cite{svyatkovskiy2020intellicode}. 
\begin{table}[h]
\small
	\centering
	\begin{tabular}{l|c|c|c|c}
		\hline
		\multirow{2}{*}{Model} & \multicolumn{2}{c|}{PY150} & \multicolumn{2}{c}{JavaCorpus} \\
		\cline{2-5}
		&EM &Edit Sim&EM &Edit Sim  \\
		\hline
		Transformer & 38.51 & 69.01& 17.00 & 50.23\\
		GPT-2 &41.73 & 70.60&27.50 & 60.36\\
		CodeGPT & 42.37 & 71.59& 30.60 & 63.45\\
		\hline
		PLBART & 38.01 & 68.46& 26.97 & 61.59\\	
		CodeT5-base & 36.97 & 67.12& 24.80 & 58.31\\
		\hline
		UniXcoder& \bf{43.12}&\bf{72.00}& \bf{32.90}&\bf{65.78}\\
		\hdashline
		\ -w/o contras& 43.02&71.94& 32.77&65.71\\
		\ -w/o cross-gen& 42.66&71.83& 32.43&65.63\\
		\hdashline
		\ -w/o comment& 42.18&71.70& 32.20&65.44\\
		\ -w/o AST& 42.56&71.87& 32.63&65.66\\
		\hdashline
		\ -using BFS& 42.83&71.85& 32.40&65.55\\
		\ -using DFS& 42.61&71.97& 32.87&65.75\\	
		\hline
	\end{tabular}
	\caption{Results of code completion task.}
	\label{table:completion_task}
\end{table}

\begin{table*}[t]
\small
	\centering
	\begin{tabular}{l|ccc|ccc|ccc|c}
		\hline
		\multirow{2}{*}{Model} & \multicolumn{3}{c|}{Ruby} & \multicolumn{3}{c|}{Python}& \multicolumn{3}{c|}{Java} & \multirow{2}{*}{Overall} \\
		\cline{2-10}
		 & Ruby & Python & Java & Ruby & Python & Java & Ruby & Python & Java &\\
		\hline
		CodeBERT & 13.55 & 3.18 & 0.71 & 3.12 & 14.39 & 0.96 & 0.55& 0.42& 7.62& 4.94 \\
		GraphCodeBERT & 17.01 & 9.29 & 6.38 & 5.01 & 19.34 & 6.92 & 1.77& 3.50& 13.31& 9.17\\
		\hline
		PLBART & 18.60 & 10.76 & 1.90 & 8.27 & 19.55 & 1.98 & 1.47& 1.27& 10.41& 8.25 \\
		CodeT5-base & 18.22 & 10.02 & 1.81 & 8.74 & 17.83 & 1.58 & 1.13& 0.81& 10.18& 7.81 \\
		\hline
        UniXcoder& \bf{29.05} & \bf{26.36} & \bf{15.16} & \bf{23.96} & \bf{30.15} & \bf{15.07} & \bf{13.61}& \bf{14.53}& \bf{16.12}& \bf{20.45} \\
        \hdashline
        \ -w/o contras& 24.03 & 17.35 & 7.12 & 15.80 & 22.52 & 7.31 & 7.55& 7.98& 13.92& 13.73 \\
        \ -w/o cross-gen& 28.73 & 24.16 & 12.92 & 21.52 & 26.66 & 12.60 & 11.14& 10.82& 13.75& 18.03 \\  
        \hdashline
        \ -w/o comment& 22.24 & 15.90 & 7.50 & 15.09 & 19.88 & 6.54 & 7.84& 7.12& 13.20& 12.81 \\
        \ -w/o AST& 27.54 & 23.37 & 10.17 & 21.75 & 27.75 & 9.94 & 9.79& 9.21& 14.06& 17.06 \\
        \hdashline
        \ -using BFS& 26.67 & 23.69 & 13.56 & 21.31 & 27.28 &13.63 & 11.90& 12.55& 14.92& 18.39 \\
        \ -using DFS& 27.13 & 22.65 & 11.62 & 20.21 & 25.92 & 11.85 & 9.59& 10.19&13.30& 16.94\\        
		\hline
	\end{tabular}
	\caption{MAP score (\%) of zero-shot setting on code-to-code search task. }
	\label{table:cross_lingual_clone_detection}
\end{table*}

In practice, the task requires a decoder-only manner to perform efficient inference. Therefore, we first compare our UniXcoder with decoder-only models (the first group) in Table \ref{table:completion_task}. As we can see, UniXcoder achieves comparable performance on both datasets and brings absolute 2.3\% gain of accuracy on java corpus, which demonstrates the effectiveness of our model for code completion. Besides, we also compare with current unified models (the second group). Since they are based the encoder-decoder framework, we fine-tune their decoders by feeding a placeholder into the encoder. Results show that UniXcoder 
outperforms PLBART and CodeT5, which demonstrates our model framework is better applied to code completion tasks.

\subsection{Zero-shot Code-to-Code Search}
\label{sec:zero-shot}
To further evaluate the performance of code fragment embeddings, we also propose a new task called zero-shot code-to-code search. Given a source code as the query, the task aims to retrieve codes with the same semantics from a collection of candidates in zero-shot setting. 
The task can help users translate from one PL to another by retrieving source codes with the same semantics. We collect 11,744/15,594/23,530 functions from the CodeNet corpus \cite{puri2021project} in Ruby/Python/Java PL. Each function solves one of 4,053 problems. We take each function as a query and retrieve all functions that solve the same problem from each 
PL. We use average MAP score as the evaluation metric. More details about
the dataset and an example can be found in Appendix \ref{appedix:zero-shot}. 

We re-implement the publicly released pre-trained models on this task using the mean vector or CLS vector of last hidden states and report the results in Table \ref{table:cross_lingual_clone_detection}. The first row is the query PL and the second row is the target PL. 
From the table, we can see that UniXcoder achieves state-of-the-art performance and about 11 points improvement on the overall score compared with GraphCodeBERT. Ablation studies further show that both multi-modal data and code fragment representation pre-training tasks can enhance UniXcoder.
 
\subsection{Model Analysis}
\paragraph{The Effect of Representation Pre-training}
We conduct ablation study to analyze the effect of code fragment representation pre-training tasks by removing contrastive learning task (\textbf{w/o constras}) and cross-modal generation task (\textbf{w/o cross-gen}). As we can see in Table \ref{table:understand_task} and \ref{table:cross_lingual_clone_detection}, two pre-training tasks significantly improve understanding tasks. Taking zero-shot code-code search task as an example, after removing contrastive learning, the performance drops from 20.45\% to 13.73\%. Besides, the two pre-training tasks also bring a small improvement on generation tasks, as shown in Table \ref{table:generation_task} and \ref{table:completion_task}. Overall, the ablation study demonstrates the effectiveness of the two pre-training tasks.

\paragraph{The Effect of Multi-modal Data}
We also study the effect of multi-modal data. By removing comment (\textbf{w/o comment}), the results from Tables indicate that code comment plays an important role in both understanding and generation tasks. For AST (\textbf{w/o AST}), we observe that injecting AST can boost the performance on all code understanding tasks. 
However, AST does not bring improvements on generation tasks, which may require a better way to incorporate AST for generation tasks. Overall, AST  and  comment  can both improve UniXcoder.

\paragraph{Comparison of Traversal Algorithms}
We compare our mapping function with other mapping functions used to map a tree into a sequence, namely BFS and DFS algorithms. As we can see, after replacing our mapping function by BFS or DFS algorithms, the performance of UniXcoder drops on both understanding and generation tasks, which demonstrates the effectiveness of our mapping function. In particular, using BFS or DFS algorithms even hurt the performance of UniXcoder on some tasks by comparing \textbf{w/o BFS (DFS)} with \textbf{w/o AST}. The main reason may be that BFS and DFS algorithms are not one-to-one mapping functions and can confuse a tree with another structure.

\begin{figure}[h]
    \centering
    \includegraphics[width=\linewidth]{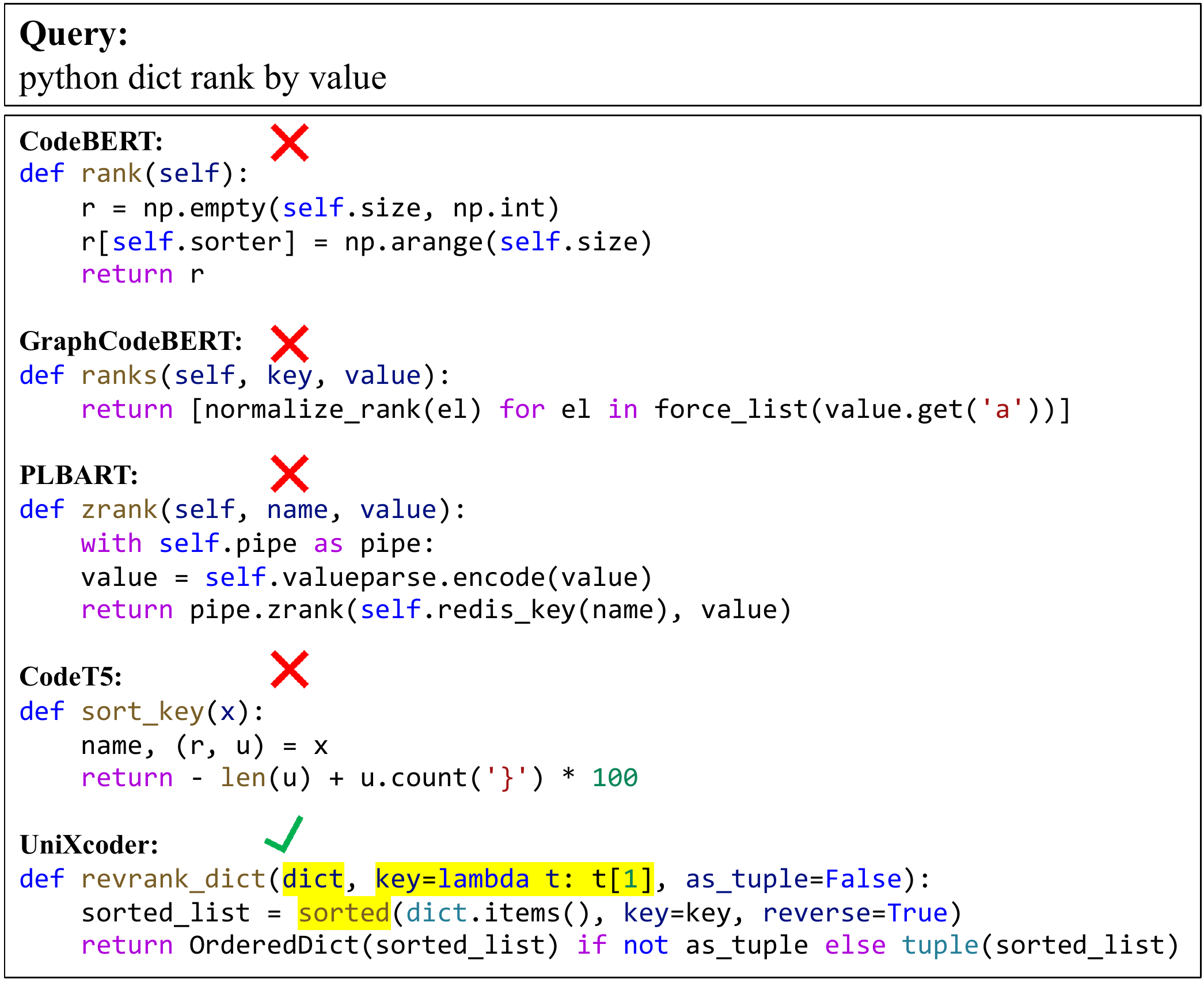}
    \caption{An examples for code search task on CosQA dataset and predictions from different models. Key clues are marked in yellow.
}  
    \label{fig:case_text_code_search}
\end{figure}

\paragraph{Case Study}
We also conduct a case study to intuitively demonstrate the effectiveness of UniXcoder, as shown in Figure~\ref{fig:case_text_code_search}. We give an example for code search task on CosQA dataset and output predictions from different models. The query \textit{``python dict rank by value''} comes from the search logs of Microsoft Bing search engine. We know that the intent of the user is to sort a dictionary by its value in Python language. 
Although the prediction from PLBART has higher lexical overlap like \textit{``rank''} and \textit{``value''}, the function is incorrect since the input of the ground truth should be a dictionary.
We can see that UniXcoder retrieves a correct function whose input is a dictionary.
Besides, although the \textit{``value''} in the query is expressed as the statement \textit{``key=lambda t: t[1]''} in the function  definition, UniXcoder can understand the code semantics and successfully retrieves the ground truth, which demonstrates the effectiveness of UniXcoder.

\section{Conclusion}
To support both code-related understanding and generation tasks, we present UniXcoder, a unified pre-trained model that incorporates semantic and syntax information from code comment and AST. We propose a one-to-one mapping method to transform AST to a sequence structure and two new pre-training tasks to learn code fragment representation.
To further investigate the performance of code representation, we propose a new downstream task of zero-shot code-to-code search and create a dataset for this task.
Experiments show that UniXcoder significantly outperforms previous works on most tasks.
Further ablation studies also show that both AST and code comment can enhance UniXcoder and reveal the effectiveness of our proposed mapping function and pre-training tasks. 

\subsubsection*{Acknowledgments}
Yanlin Wang is the corresponding author. Daya Guo and Jian Yin are supported by the National Natural Science Foundation of China (U1811264, U1811262, U1811261, U1911203, U2001211), Guangdong Basic and Applied Basic Research Foundation (2019B1515130001), Key-Area Research and Development Program of Guangdong Province (2018B010107005, 2020B0101100001).
\bibliography{anthology,custom}

\begin{thebibliography}{31}
\expandafter\ifx\csname natexlab\endcsname\relax\def\natexlab#1{#1}\fi

\bibitem[{Ahmad et~al.(2021)Ahmad, Chakraborty, Ray, and
  Chang}]{ahmad2021unified}
Wasi Ahmad, Saikat Chakraborty, Baishakhi Ray, and Kai-Wei Chang. 2021.
\newblock Unified pre-training for program understanding and generation.
\newblock In \emph{Proceedings of the 2021 Conference of the North American
  Chapter of the Association for Computational Linguistics: Human Language
  Technologies}, pages 2655--2668.

\bibitem[{Allamanis and Sutton(2013)}]{allamanis2013mining}
Miltiadis Allamanis and Charles Sutton. 2013.
\newblock Mining source code repositories at massive scale using language
  modeling.
\newblock In \emph{2013 10th Working Conference on Mining Software Repositories
  (MSR)}, pages 207--216. IEEE.

\bibitem[{Brown et~al.(2020)Brown, Mann, Ryder, Subbiah, Kaplan, Dhariwal,
  Neelakantan, Shyam, Sastry, Askell et~al.}]{brown2020language}
Tom~B Brown, Benjamin Mann, Nick Ryder, Melanie Subbiah, Jared Kaplan, Prafulla
  Dhariwal, Arvind Neelakantan, Pranav Shyam, Girish Sastry, Amanda Askell,
  et~al. 2020.
\newblock Language models are few-shot learners.
\newblock \emph{arXiv preprint arXiv:2005.14165}.

\bibitem[{Buratti et~al.(2020)Buratti, Pujar, Bornea, McCarley, Zheng,
  Rossiello, Morari, Laredo, Thost, Zhuang et~al.}]{buratti2020exploring}
Luca Buratti, Saurabh Pujar, Mihaela Bornea, Scott McCarley, Yunhui Zheng,
  Gaetano Rossiello, Alessandro Morari, Jim Laredo, Veronika Thost, Yufan
  Zhuang, et~al. 2020.
\newblock Exploring software naturalness throughneural language models.
\newblock \emph{arXiv preprint arXiv:2006.12641}.

\bibitem[{Devlin et~al.(2018)Devlin, Chang, Lee, and
  Toutanova}]{devlin2018bert}
Jacob Devlin, Ming-Wei Chang, Kenton Lee, and Kristina Toutanova. 2018.
\newblock Bert: Pre-training of deep bidirectional transformers for language
  understanding.
\newblock \emph{arXiv preprint arXiv:1810.04805}.

\bibitem[{Dong et~al.(2019)Dong, Yang, Wang, Wei, Liu, Wang, Gao, Zhou, and
  Hon}]{dong2019unified}
Li~Dong, Nan Yang, Wenhui Wang, Furu Wei, Xiaodong Liu, Yu~Wang, Jianfeng Gao,
  Ming Zhou, and Hsiao-Wuen Hon. 2019.
\newblock Unified language model pre-training for natural language
  understanding and generation.
\newblock \emph{arXiv preprint arXiv:1905.03197}.

\bibitem[{Feng et~al.(2020)Feng, Guo, Tang, Duan, Feng, Gong, Shou, Qin, Liu,
  Jiang et~al.}]{feng2020codebert}
Zhangyin Feng, Daya Guo, Duyu Tang, Nan Duan, Xiaocheng Feng, Ming Gong, Linjun
  Shou, Bing Qin, Ting Liu, Daxin Jiang, et~al. 2020.
\newblock Codebert: A pre-trained model for programming and natural languages.
\newblock \emph{arXiv preprint arXiv:2002.08155}.

\bibitem[{Gao et~al.(2021)Gao, Yao, and Chen}]{gao2021simcse}
Tianyu Gao, Xingcheng Yao, and Danqi Chen. 2021.
\newblock Simcse: Simple contrastive learning of sentence embeddings.
\newblock \emph{arXiv preprint arXiv:2104.08821}.

\bibitem[{Guo et~al.(2020)Guo, Ren, Lu, Feng, Tang, Shujie, Zhou, Duan,
  Svyatkovskiy, Fu et~al.}]{guo2020graphcodebert}
Daya Guo, Shuo Ren, Shuai Lu, Zhangyin Feng, Duyu Tang, LIU Shujie, Long Zhou,
  Nan Duan, Alexey Svyatkovskiy, Shengyu Fu, et~al. 2020.
\newblock Graphcodebert: Pre-training code representations with data flow.
\newblock In \emph{International Conference on Learning Representations}.

\bibitem[{Huang et~al.(2021)Huang, Tang, Shou, Gong, Xu, Jiang, Zhou, and
  Duan}]{huang2021cosqa}
Junjie Huang, Duyu Tang, Linjun Shou, Ming Gong, Ke~Xu, Daxin Jiang, Ming Zhou,
  and Nan Duan. 2021.
\newblock Cosqa: 20,000+ web queries for code search and question answering.
\newblock \emph{arXiv preprint arXiv:2105.13239}.

\bibitem[{Husain et~al.(2019)Husain, Wu, Gazit, Allamanis, and
  Brockschmidt}]{husain2019codesearchnet}
Hamel Husain, Ho-Hsiang Wu, Tiferet Gazit, Miltiadis Allamanis, and Marc
  Brockschmidt. 2019.
\newblock Codesearchnet challenge: Evaluating the state of semantic code
  search.
\newblock \emph{arXiv preprint arXiv:1909.09436}.

\bibitem[{Iyer et~al.(2018)Iyer, Konstas, Cheung, and
  Zettlemoyer}]{iyer2018mapping}
Srinivasan Iyer, Ioannis Konstas, Alvin Cheung, and Luke Zettlemoyer. 2018.
\newblock Mapping language to code in programmatic context.
\newblock In \emph{Proceedings of the 2018 Conference on Empirical Methods in
  Natural Language Processing}, pages 1643--1652.

\bibitem[{Jiang et~al.(2021)Jiang, Zheng, Lyu, Li, and Lyu}]{jiang2021treebert}
Xue Jiang, Zhuoran Zheng, Chen Lyu, Liang Li, and Lei Lyu. 2021.
\newblock Treebert: A tree-based pre-trained model for programming language.
\newblock \emph{arXiv preprint arXiv:2105.12485}.

\bibitem[{Kanade et~al.(2019)Kanade, Maniatis, Balakrishnan, and
  Shi}]{kanade2019pre}
Aditya Kanade, Petros Maniatis, Gogul Balakrishnan, and Kensen Shi. 2019.
\newblock Pre-trained contextual embedding of source code.
\newblock \emph{arXiv preprint arXiv:2001.00059}.

\bibitem[{Lewis et~al.(2019)Lewis, Liu, Goyal, Ghazvininejad, Mohamed, Levy,
  Stoyanov, and Zettlemoyer}]{lewis2019bart}
Mike Lewis, Yinhan Liu, Naman Goyal, Marjan Ghazvininejad, Abdelrahman Mohamed,
  Omer Levy, Ves Stoyanov, and Luke Zettlemoyer. 2019.
\newblock Bart: Denoising sequence-to-sequence pre-training for natural
  language generation, translation, and comprehension.
\newblock \emph{arXiv preprint arXiv:1910.13461}.

\bibitem[{Lin and Och(2004)}]{lin2004orange}
Chin-Yew Lin and Franz~Josef Och. 2004.
\newblock Orange: a method for evaluating automatic evaluation metrics for
  machine translation.
\newblock In \emph{COLING 2004: Proceedings of the 20th International
  Conference on Computational Linguistics}, pages 501--507.

\bibitem[{Liu et~al.(2020)Liu, Li, Zhao, and Jin}]{liu2020multi}
Fang Liu, Ge~Li, Yunfei Zhao, and Zhi Jin. 2020.
\newblock Multi-task learning based pre-trained language model for code
  completion.
\newblock In \emph{Proceedings of the 35th IEEE/ACM International Conference on
  Automated Software Engineering}, pages 473--485.

\bibitem[{Liu et~al.(2019)Liu, Ott, Goyal, Du, Joshi, Chen, Levy, Lewis,
  Zettlemoyer, and Stoyanov}]{liu2019roberta}
Yinhan Liu, Myle Ott, Naman Goyal, Jingfei Du, Mandar Joshi, Danqi Chen, Omer
  Levy, Mike Lewis, Luke Zettlemoyer, and Veselin Stoyanov. 2019.
\newblock Roberta: A robustly optimized bert pretraining approach.
\newblock \emph{arXiv preprint arXiv:1907.11692}.

\bibitem[{Lu et~al.(2021)Lu, Guo, Ren, Huang, Svyatkovskiy, Blanco, Clement,
  Drain, Jiang, Tang et~al.}]{lu2021codexglue}
Shuai Lu, Daya Guo, Shuo Ren, Junjie Huang, Alexey Svyatkovskiy, Ambrosio
  Blanco, Colin Clement, Dawn Drain, Daxin Jiang, Duyu Tang, et~al. 2021.
\newblock Codexglue: A machine learning benchmark dataset for code
  understanding and generation.
\newblock \emph{arXiv preprint arXiv:2102.04664}.

\bibitem[{Mou et~al.(2016)Mou, Li, Zhang, Wang, and Jin}]{mou2016convolutional}
Lili Mou, Ge~Li, Lu~Zhang, Tao Wang, and Zhi Jin. 2016.
\newblock Convolutional neural networks over tree structures for programming
  language processing.
\newblock In \emph{Proceedings of the Thirtieth AAAI Conference on Artificial
  Intelligence}, pages 1287--1293.

\bibitem[{Puri et~al.(2021)Puri, Kung, Janssen, Zhang, Domeniconi, Zolotov,
  Dolby, Chen, Choudhury, Decker et~al.}]{puri2021project}
Ruchir Puri, David~S Kung, Geert Janssen, Wei Zhang, Giacomo Domeniconi,
  Vladmir Zolotov, Julian Dolby, Jie Chen, Mihir Choudhury, Lindsey Decker,
  et~al. 2021.
\newblock Project codenet: A large-scale ai for code dataset for learning a
  diversity of coding tasks.
\newblock \emph{arXiv preprint arXiv:2105.12655}.

\bibitem[{Radford et~al.(2018)Radford, Narasimhan, Salimans, and
  Sutskever}]{radford2018improving}
Alec Radford, Karthik Narasimhan, Tim Salimans, and Ilya Sutskever. 2018.
\newblock Improving language understanding by generative pre-training.
\newblock \emph{URL https://s3-us-west-2. amazonaws.
  com/openai-assets/researchcovers/languageunsupervised/language understanding
  paper. pdf}.

\bibitem[{Radford et~al.(2019)Radford, Wu, Child, Luan, Amodei, Sutskever
  et~al.}]{radford2019language}
Alec Radford, Jeffrey Wu, Rewon Child, David Luan, Dario Amodei, Ilya
  Sutskever, et~al. 2019.
\newblock Language models are unsupervised multitask learners.
\newblock \emph{OpenAI blog}, 1(8):9.

\bibitem[{Raffel et~al.(2019)Raffel, Shazeer, Roberts, Lee, Narang, Matena,
  Zhou, Li, and Liu}]{raffel2019exploring}
Colin Raffel, Noam Shazeer, Adam Roberts, Katherine Lee, Sharan Narang, Michael
  Matena, Yanqi Zhou, Wei Li, and Peter~J Liu. 2019.
\newblock Exploring the limits of transfer learning with a unified text-to-text
  transformer.
\newblock \emph{arXiv preprint arXiv:1910.10683}.

\bibitem[{Raychev et~al.(2016)Raychev, Bielik, and
  Vechev}]{raychev2016probabilistic}
Veselin Raychev, Pavol Bielik, and Martin Vechev. 2016.
\newblock Probabilistic model for code with decision trees.
\newblock \emph{ACM SIGPLAN Notices}, pages 731--747.

\bibitem[{Sennrich et~al.(2015)Sennrich, Haddow, and
  Birch}]{sennrich2015neural}
Rico Sennrich, Barry Haddow, and Alexandra Birch. 2015.
\newblock Neural machine translation of rare words with subword units.
\newblock \emph{arXiv preprint arXiv:1508.07909}.

\bibitem[{Svajlenko et~al.(2014)Svajlenko, Islam, Keivanloo, Roy, and
  Mia}]{svajlenko2014towards}
Jeffrey Svajlenko, Judith~F Islam, Iman Keivanloo, Chanchal~K Roy, and
  Mohammad~Mamun Mia. 2014.
\newblock Towards a big data curated benchmark of inter-project code clones.
\newblock In \emph{2014 IEEE International Conference on Software Maintenance
  and Evolution}, pages 476--480. IEEE.

\bibitem[{Svyatkovskiy et~al.(2020)Svyatkovskiy, Deng, Fu, and
  Sundaresan}]{svyatkovskiy2020intellicode}
Alexey Svyatkovskiy, Shao~Kun Deng, Shengyu Fu, and Neel Sundaresan. 2020.
\newblock Intellicode compose: Code generation using transformer.
\newblock \emph{arXiv preprint arXiv:2005.08025}.

\bibitem[{Vaswani et~al.(2017)Vaswani, Shazeer, Parmar, Uszkoreit, Jones,
  Gomez, Kaiser, and Polosukhin}]{vaswani2017attention}
Ashish Vaswani, Noam Shazeer, Niki Parmar, Jakob Uszkoreit, Llion Jones,
  Aidan~N Gomez, {\L}ukasz Kaiser, and Illia Polosukhin. 2017.
\newblock Attention is all you need.
\newblock In \emph{Advances in neural information processing systems}, pages
  5998--6008.

\bibitem[{Wang et~al.(2022)Wang, Yasheng~Wang, Zhou, Wan, Liu, Li, Wu, Liu, and
  Jiang}]{wang2022syncobert}
Xin Wang, Fei~Mi Yasheng~Wang, Pingyi Zhou, Yao Wan, Xiao Liu, Li~Li, Hao Wu,
  Jin Liu, and Xin Jiang. 2022.
\newblock Syncobert: Syntax-guided multi-modal contrastive pre-training for
  code representation.

\bibitem[{Wang et~al.(2021)Wang, Wang, Joty, and Hoi}]{wang2021codet5}
Yue Wang, Weishi Wang, Shafiq Joty, and Steven~CH Hoi. 2021.
\newblock Codet5: Identifier-aware unified pre-trained encoder-decoder models
  for code understanding and generation.
\newblock \emph{arXiv preprint arXiv:2109.00859}.

\end{thebibliography}
\bibliographystyle{acl_natbib}

\appendix

\section{Proof for Mapping Function}
\label{sec:proof}
In this section, we show that the function $\mathcal{F}$ described in Algorithm \ref{alg:mapping} is a one-to-one mapping function. using a \textbf{proof by induction}. 

\paragraph{Lemma 1:}
Given a tree $\mathcal{T}$ and the mapped sequence $\mathcal{F}(\mathcal{T})=\{x_1,x_2,...,x_m\}$, the first element $x_0$ is the root of $\mathcal{T}$ with a \textit{left} suffix and the last element $x_m$ is the root of $\mathcal{T}$ with a \textit{right} suffix.

\paragraph{Lemma 2:}
An internal node only occurs twice in the mapped sequence $\mathcal{F}(\mathcal{T})$. One is with \textit{left} suffix, and the other is with \textit{right} suffix. Therefore, the sequence has the same number of elements with \textit{left} suffix and \textit{right} suffix. 

\paragraph{Lemma 3:}
Since a node with a \textit{left} suffix occurs before the same node with a \textit{right} suffix (see line 7 and 11 of the Algorithm), an element $x_i$ with  a \textit{right} suffix must match another element $x_j$ with a \textit{left} suffix (i.e. coming from the same node in the tree) in the left side, i.e. $j<i$. 

\paragraph{Proof:} 
In order to prove $\mathcal{F}$ is a one-to-one function, given two trees $\mathcal{T}_1$ and $\mathcal{T}_2$, we need to prove that $\mathcal{F}(\mathcal{T}_1)\neq\mathcal{F}(\mathcal{T}_2)$ if $\mathcal{T}_1\neq\mathcal{T}_2$. For easier proof, we prove its equivalent contrapositive statement, i.e. $\mathcal{T}_1=\mathcal{T}_2$  if  $\mathcal{F}(\mathcal{T}_1)=\mathcal{F}(\mathcal{T}_2)$.

\paragraph{Base case:} 
When the depth $h$ of $\mathcal{T}_1$ is 1, $\mathcal{T}_1$ only has a node $r$ and $\mathcal{F}(\mathcal{T}_1)=\{r\}$. Since $\mathcal{F}(\mathcal{T}_2)=\mathcal{F}(\mathcal{T}_1)$, $\mathcal{T}_2$ contains one node, otherwise the length of $\mathcal{F}(\mathcal{T}_2)$ will be more than 2. Since  $\mathcal{F}(\mathcal{T}_2)=\{r\}$, the root of $\mathcal{T}_2$ is also $r$. Therefore, $\mathcal{T}_1=\mathcal{T}_2$ and $\mathcal{F}$ is a one-to-one function for $h=1$.

\paragraph{Inductive hypothesis:} 
When $h = 2,3,..,n$, suppose $\mathcal{F}$ is a one-to-one function.

\paragraph{Inductive step:} 
Now, we prove that the hypothesis is true for $h=n+1\ge2$.

We let $\mathcal{F}(\mathcal{T}_1)=\{x_1,x_2,...,x_m\}$, where $x_j$ is a leaf or a node with a \textit{left} (or \textit{right}) suffix. 
Since $\mathcal{F}(\mathcal{T}_1)=\mathcal{F}(\mathcal{T}_2)$, the first elements of two mapped sequences are same as each other. According to Lemma 1, $\mathcal{T}_1$ and $\mathcal{T}_2$ have the same root.

Let the leftmost subtree of $\mathcal{T}_1$ and $\mathcal{T}_2$ as $\mathcal{T}_{s_1}$  and $\mathcal{T}_{s_2}$, respectively. We prove $\mathcal{T}_{s_1}=\mathcal{T}_{s_2}$ now. According to the Algorithm, we know that $\mathcal{F}(\mathcal{T}_{s_1})$ and $\mathcal{F}(\mathcal{T}_{s_2})$ start with $x_2$ and end with one element. Suppose $\mathcal{F}(\mathcal{T}_{s_1})=\{x_2,..,x_i\}$ and $\mathcal{F}(\mathcal{T}_{s_2})=\{x_2,..,x_j\}$.  
According to Lemma 1 and 2, $S=\{x_3,..,x_i\}$ has one more element with a \textit{right} suffix. Therefore, $x_0$ must match one element $x_k$ ($3\le k \le i$) in $S$, otherwise there will be an element with a \textit{right} suffix  that cannot match any element.
If $i\neq j$ (suppose $j>i$), $x_0$ will match $x_j$ according to Lemma 1. 
However, the root node occurs three times $x_0$, $x_k$ and $x_j$, which will contradict Lemma 2.
Therefore, we get that $i=j$ and $\mathcal{F}(\mathcal{T}_{s_1})=\mathcal{F}(\mathcal{T}_{s_2})$.
According to the hypothesis, we get that $\mathcal{T}_{s_1}=\mathcal{T}_{s_2}$, since the depth of $\mathcal{F}(\mathcal{T}_{s_1})$ is less than $n+1$. In the same way, it can be proved that other subtrees of $\mathcal{T}_1$ and $\mathcal{T}_2$ are also the same. Thus, we get that $\mathcal{T}_1=\mathcal{T}_2$.

\paragraph{Conclusion:}
By the principle of induction, it follows that the hypothesis is true for all $h\ge 2$ and our mapping function is one-to-one.



\begin{table*}[t]
\small
	\centering
	\begin{tabular}{lccccccc}
		\hline
		Model & Ruby & Javascript & Go & Python & Java & Php & Overall\\
		\hline
		RoBERTa & 58.7 & 51.7 & 85.0 & 58.7 & 59.9 & 56.0 & 61.7 \\
		CodeBERT & 67.9 & 62.0 & 88.2 & 67.2 & 67.6 & 62.8 & 69.3 \\
		GraphCodeBERT  & 70.3 & 64.4 & 89.7 & 69.2 & 69.1 & 64.9 & 71.3 \\
        \textsc{SynCoBERT} & 72.2 & 67.7 & 91.3 & \bf{72.4} & 72.3 & \bf{67.8} & 74.0 \\	
		\hline
		PLBART  & 67.5 & 61.6 & 88.7 & 66.3 & 66.3 & 61.1 & 68.5 \\
		CodeT5-base  & 71.9 & 65.5 & 88.8 & 69.8 & 68.6 & 64.5 & 71.5\\	

		\hline
        UniXcoder & \bf{74.0} & \bf{68.4} & \bf{91.5}& 72.0 &\bf{72.6} & 67.6&\bf{74.4}  \\
        \hdashline
        \ -w/o contras&73.4&67.0&91.3&71.3&71.7&66.7&73.6 \\
        \ -w/o cross-gen & 73.0 & 67.8 & 91.3& 71.9 &72.4 & 67.3&74.0 \\
        \hdashline
        \ -w/o comment & 72.0 & 65.7 & 91.1& 70.4 &70.6 & 65.5&72.6 \\
        \ -w/o AST & 73.8 & 68.0 & 91.4& 72.3 &72.3 & 67.4&74.2 \\
        \hdashline
        \ -using BFS & 73.4&68.2 & 91.3&72.2&72.2&67.3&74.1\\
        \ -using DFS & 73.5&68.3 & 91.2&72.3&72.1&67.6&74.2\\        
	    \hline
	\end{tabular}
	\caption{Results of code search task over six programming languages. }
	\label{table:codesearch}    
\end{table*}

\section{Pre-training Setting}
\label{appedix:pre-training-setting}
UniXcoder uses 12 layers of Transformer with 768 dimensional hidden states and 12 attention heads. 
We follow \citet{liu2019roberta} to train a byte-pair encoding vocabulary \cite{sennrich2015neural} with 50K subword units for programming languages and add 1,416 additional special tokens into the vocabulary to represent non-terminal symbols in AST. The pre-training multi-modal data we use includes 2.3M functions paired with comments from CodeSearchNet dataset  \cite{husain2019codesearchnet} for six programming languages (i.e. ruby, java, python, php, go and javascript). We leverage tree-sitter\footnote{\url{https://github.com/tree-sitter/tree-sitter}} as the parser to extract AST from PL.

We pre-train the model on 4 DGX-2 machines, each having 16 NVIDIA Tesla V100 with 32GB memory. During pre-training, we set both the max length of input sequence and batch size as 1024, and use the Adam optimizer to update model parameters with 2e-4 learning rate.  As proven in \citet{feng2020codebert}, unimodal data like text is also useful for code-related downstream tasks. Therefore, we first pre-train our UniXcoder with MLM, ULM and denoising objective on C4 dataset \cite{raffel2019exploring} and 4.1M unimodal code from CodeSearchNet for 500k and 200k steps, respectively. We further pre-train on the multi-modal data with all pre-training objectives for 100k steps. The total time for pre-training UniXcoder is about 8 days.
At each iteration, we alternate each objective to pre-train the model and follow \citet{guo2020graphcodebert} to sample each batch from the same programming language according to a distribution ${\{q_i\}}_{i=1...N}$ as Equation \ref{sample}, where $n_i$ is number of examples for $i$-th programming language and $\alpha$=0.7. Sampling with this distribution could alleviates the bias towards high-resource languages.
\begin{equation}\label{sample}
q_i=\frac{p^\alpha_i}{\sum_{N}^{j=1}p^\alpha_j},\ \ p_i=\frac{n_i}{\sum_{N}^{k=1}n_k}
\end{equation}

\section{Fine-tuning Setting}
\label{appedix:fine-tuning-setting}
\subsection{Clone Detection}
Clone detection aims to measure the similarity between two code fragments. We conduct experiments on POJ-104 \cite{mou2016convolutional} and BigCloneBench \citep{svajlenko2014towards} datasets. 

For POJ-104 dataset, it consists of 104 problems and includes 500 C/C++ programs for each problem. The datasets are splited into 64/16/24 problems for training, validation, and testing, and the task aims to retrieve other programs that solve the same problem given a program. The probability of true clone is calculated by cosine similarity between two mean vectors of last hidden states of UniXcoder. We set the learning rate as 2e-5, the batch size as 8, and the max sequence length as 400. We use the Adam optimizer to fine-tune the model for 2 epochs.

For BigCloneBench dataset, we use the dataset provided by \citet{lu2021codexglue}, which includes 901,724/416,328/416,328 examples from 10 different functionalities for training/validation/testing. Following previous works, we also treat the task as a binary classification to fine-tune UniXcoder. The true clone probability of two inputs is calculated by cosine similarity between the mean vectors of last hidden states.  In the fine-turning step, we set the learning rate as 5e-5, the batch size as 16, and the max sequence length as 512. We update model parameters using the Adam optimizer and perform early stopping on the development set.
\subsection{Code Search}
Code search aims to search the most relevant code from a collection of candidates given a natural language query. We conduct experiments on three datasets, namely CSN \cite{guo2020graphcodebert}, AdvTest \cite{lu2021codexglue} and CosQA \cite{huang2021cosqa}. 

\begin{table}[h]
\small
	\begin{center}
		\begin{tabular}{l|c|c|c|c}
			\hline 
			Language & Training& Dev& Testing & Candidates \\
			\hline 
			Go&167,288&7,325&8,122&28,120\\
			Java&164,923&5,183&10,955&40,347\\
			JavaScript&58,025&3,885&3,291&13,981\\
			PHP&241,241&12,982&14,014&52,660\\
			Python&251,820&13,914&14,918&43,827\\
			Ruby&24,927&1,400&1,261&4,360\\
			\hline 
		\end{tabular}
	\caption{Data statistics about CSN dataset provided by \citet{guo2020graphcodebert}. \textbf{Training}/\textbf{Dev}/\textbf{Testing} means the number of query for training/validation/testing dataset.}
	\label{table-codesearchnet-data-statistic}
	\end{center}
\end{table}
\begin{table*}[t]
\small
	\centering
	\begin{tabular}{lccccccc}
		\hline
		Model & Ruby & Javascript & Go & Python & Java & Php & Overall\\
		\hline
		RoBERTa & 11.70 & 11.90 & 17.72 & 18.14 & 16.47 & 24.02 & 16.57 \\
		CodeBERT & 12.16 & 14.90 & 18.07 & 19.06 & 17.65 & 25.16 & 17.83 \\
		GraphCodeBERT  & 12.39 & 14.81 & 18.41 & 18.06 & 19.00 & 25.59 & 18.04 \\
        \hline
		PLBART  & 14.11 & 15.56 & 18.91 & 19.30 & 18.45 & 23.58 & 18.32 \\
		CodeT5-base  & \bf{15.24} & \bf{16.16} & \bf{19.56} & \bf{20.01} & 20.31 & 26.03 & \bf{19.55}\\	
		\hline
        UniXcoder&14.87&15.85&19.07&19.13&20.31&26.54&19.30 \\	
        \hdashline
        \ -w/o contras&14.72&15.41&19.16&19.01&20.30&26.60&19.20 \\
        \ -w/o cross-gen&14.90&15.96&18.60&19.06&\bf{20.50}&26.62&19.27 \\
        \hdashline
        \ -w/o
        comment&14.25&15.50&18.80&18.83&20.25&26.17&18.97 \\
        \ -w/o AST&15.09&15.97&19.04&19.16&20.07&\bf{26.67}&19.33 \\
        \hdashline
        \ -using BFS & 14.74&15.69 & 18.97&19.03&20.58&26.45&19.24\\
        \ -using DFS & 14.81&15.88 & 18.98&19.15&20.26&26.40&19.25\\        
	    \hline
	\end{tabular}
	\caption{Results of code summarization task over six programming languages. }
	\label{table:code_sum}
\end{table*}
For CSN dataset, it  is  constructed  from  CodeSearchNet dataset for six languages but filter lowquality queries by handcrafted rules. We list data statistics about the dataset in Table \ref{table-codesearchnet-data-statistic}. We set the learning rate as 2e-5, the batch size as 64, and the max sequence length of PL and NL as 256 and 128, respectively. We use the Adam optimizer to fine-tune the model for 10 epochs and perform early stopping on the development set. In Table \ref{table:codesearch}, we also give more detailed results of different models for each programming language.

For AdvTest dataset, it comes form Python language of CSN dataset but \citet{lu2021codexglue} normalizes python function and variable names to better test the understanding and generalization abilities of models. We use the same hyper-parameters as CSN dataset but fine-tune the model for 2 epochs.

For CosQA dataset, \citet{huang2021cosqa} use 20,604 search logs of the Microsoft Bing search engine as queries and each log is annotated by at least 3 human annotators. We use the same hyper-parameters as CSN dataset but fine-tune the model.

For the three datasets, we all use cosine similarity between two mean vectors of last hidden states as relevant scores and take other vectors in the same batch as negative examples.

\subsection{Code Summarization}
Code summarization aims to generate an NL summary of a code snippet. We use the dataset provided by CodeXGLUE team \cite{lu2021codexglue} for this task. The dataset includes six programming languages, including Ruby, JavaScript, Go, Python, Java, and PHP. We list data statistics about the dataset in Table \ref{table-summ-data-statistic}. We set the learning rate as 5e-5, the batch size as 48, and the max sequence length of source and target as 256 and 128, respectively. We use the Adam optimizer to fine-tune the model for 10 epochs and perform early stopping on the development set. For inference, we set beam size as 10. In Table \ref{table:code_sum}, we also give more detailed results of different models for each programming language.

\begin{table}[h]
\small
\begin{center}
\begin{tabular}{lccc}
	\hline
	Language & Training & Dev & Testing \\
	\hline
	Go&167,288&7,325&8,122\\
	Java&164,923&5,183&10,955\\
	JavaScript&58,025&3,885&3,291\\
	PHP&241,241&12,982&14,014\\
	Python&251,820&13,914&14,918\\
	Ruby&24,927&1,400&1,261\\
	\hline
\end{tabular}
\caption{Data statistics about the dataset for the code summarization task.}
\label{table-summ-data-statistic}
\end{center}
\end{table}

\subsection{Code Generation}
Code generation aims to generate a code snippet based on an NL description. We use CONCODE \cite{iyer2018mapping} dataset, which is collected from about 33k Java projects on GitHub. It contains 100k/2k/2k examples for training/validation/testing. Each example consists of an NL description, code environments and code snippets. The environment is provided by the rest of the class, including member variables and member functions in the class.  We set the learning rate as 5e-5, the batch size as 32, and the max sequence length of source and target as 350 and 150, respectively. We use the Adam optimizer to fine-tune the model for 30 epochs and perform early stopping on the development set. For inference, we set beam size as 3. 

\begin{figure*}[t]
    \centering
    \includegraphics[width=0.98\linewidth]{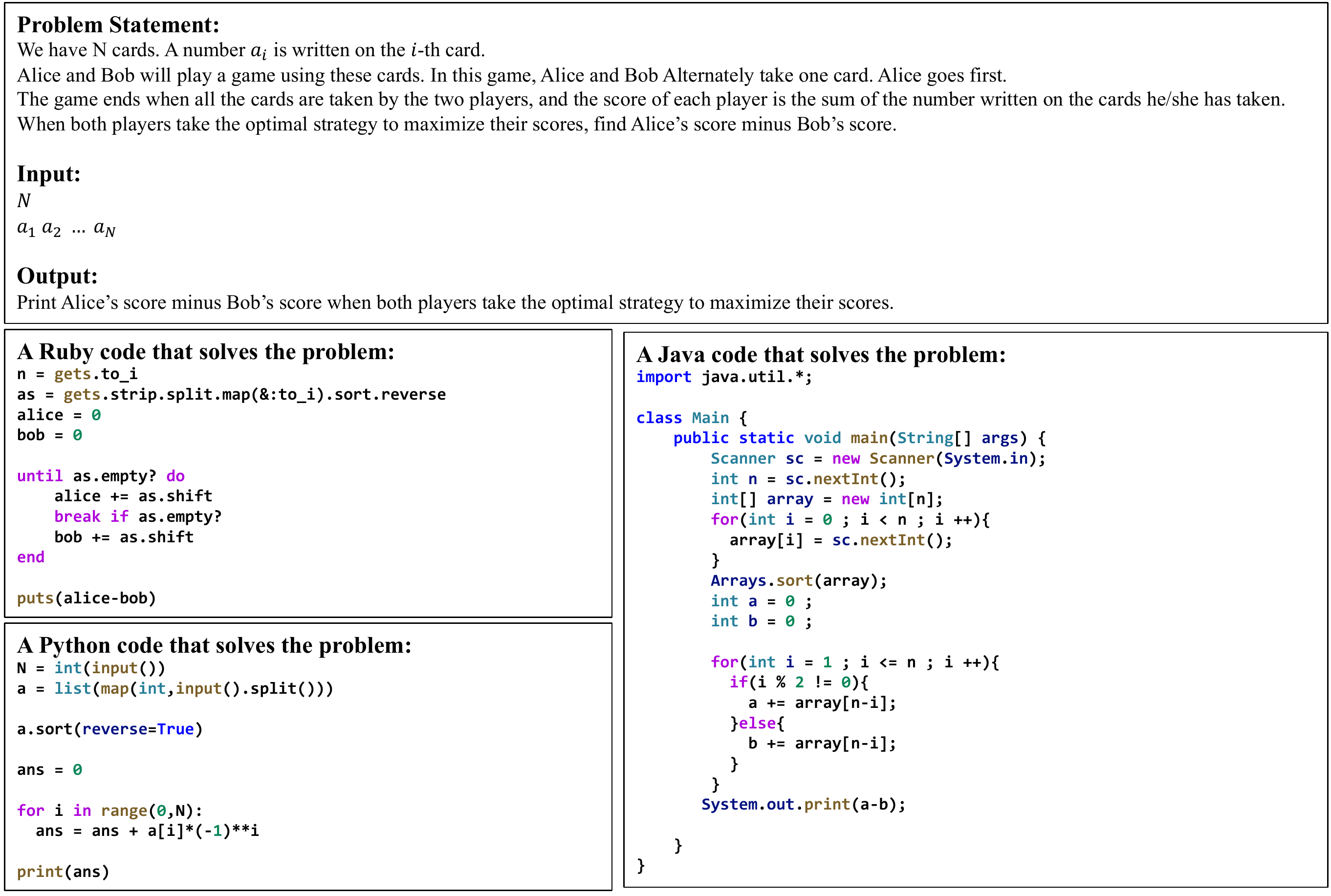}
    \caption{ An example for zero-shot code-to-code search. Three codes for Ruby, Python and Java all solve the same problem mentioned in the Figure. Therefore, they have same semantics in different programming languages.
}  
    \label{fig:case_code_code_search}
\end{figure*}
\subsection{Code Completion}
In this paper, we mainly focus on line-level code completion. We use \textbf{PY150} \cite{raychev2016probabilistic} and \textbf{Github Java Corpus} \cite{allamanis2013mining} provided by CodeXGLUE \cite{lu2021codexglue}. 

\textbf{PY150} is a Python dataset \cite{raychev2016probabilistic} containing 150,000 Python source files collected from Github. \citet{lu2021codexglue} create 10,000 examples from different files in the test set of PY150 for testing and select lines to be predicted at random. The average number of tokens in input and output are 489.11 and 6.56, respectively. 

\textbf{Github Java Corpus} is collected by \citet{allamanis2013mining} over 14 thousand Java projects from Github. \citet{lu2021codexglue} create 3,000 examples for testing from different files in the test set of the corpus. The average numbers of tokens are 350.62 and 10.49 in input and output, respectively.

For two datasets, we both follow \citet{lu2021codexglue} to use the same CodeSearchNet dataset to fine-tune UniXcoder for 10 epochs. We set the learning rate for \textbf{PY150} as 2e-4 and for \textbf{Java Corpus} as 2e-5. The batch size is 32 and the max sequence length is 1024. For inference, we set beam size as 5. 

\subsection{Zero-shot Code-to-Code Search}
\label{appedix:zero-shot}
To evaluate the performance of code fragment embeddings, we propose a new task, called zero-shot code-to-code search. Given a source code as the query, the task aims to retrieve codes with the same semantics from a collection of candidates in zero-shot setting. We give an example in Figure \ref{fig:case_code_code_search}.

We collect 11,744/15,594/23,530 functions from CodeNet corpus \cite{puri2021project} for Ruby/Python/Java PL. Each function solves one of 4,053 problems. The task is to take each function as the query and retrieve functions that solves the same problem from each PL. In zero-shot testing, we set the max sequence length as 512 and use cosine similarity between two mean vectors of last hidden states as relevant scores. We then sort the candidates by the scores to calculate MAP score.

\end{document}